\documentclass[cameraready]{Interspeech}

\title{Easper: An Accessible ASR Pipeline for Language Documentation}

\author[affiliation={1}, orcid=0000-0003-2155-4101,correspondingauthor]{Aso}{Mahmudi}
\author[affiliation={1}, orcid=0000-0003-3806-1493]{Ting}{Dang}
\author[affiliation={1}, orcid=0000-0002-4058-5459]{Ekaterina}{Vylomova}
\author[affiliation={2}, orcid=0000-0001-8797-1018]{Nick}{Thieberger}

\address{
    $^1$ School of Computing and Information Systems, The University of Melbourne, Australia \\
    $^2$ School of Languages and Linguistics, The University of Melbourne, Australia 
}

\email{amahmudi@student.unimelb.edu.au}


\usepackage{comment}
\usepackage{graphicx}   
\usepackage{subcaption} 
\usepackage{soul, xcolor}

\begin{document}

\maketitle

\begin{abstract}
Audio transcription is a critical bottleneck in language documentation. While multilingual Automatic Speech Recognition (ASR) models like Whisper offer solutions, field linguists often lack the expertise to utilise them. We present Easper, an open-source, no-code workflow enabling linguists to iteratively fine-tune ASR models via cloud resources directly from ELAN annotations. Deploying ASR also raises a cold start problem: deciding which recordings to transcribe first to bootstrap an accurate model. Using Easper, we evaluate transcription prioritisation strategies on three Vanuatu languages (Bislama, Nafsan, Nguna). We fine-tune models by recording session, comparing Character Error Rate trajectories when prioritising acoustic cleanliness versus linguistic richness. We demonstrate that prioritising lexically rich narratives and increasing acoustic-phonetic repetition, even in noisy environments, leads to faster improvements in transcription quality.

\end{abstract}

\section{Introduction}
Documenting endangered languages is a race against time. Field linguists collect vast amounts of audio recordings to capture linguistic diversity, yet the transcription of this data remains a critical bottleneck \cite{foley2018building}. It requires careful listening, speaker segmentation, and consistent orthographic choices across hours of recordings \cite{seifart2018language}. To tackle this bottleneck, researchers have proposed shifting the documentation paradigm. For example, Bird \cite{bird-2020-sparse} argued for ``sparse transcription,'' where annotations link segments of speech to selected meaningful units rather than providing exhaustive word-level text.

While sparse methods are valuable, producing full transcripts remains a primary goal for creating accessible community resources, searchable archives, and robust linguistic analyses. Automatic speech recognition (ASR) systems have the potential to significantly accelerate the transcription of audio recordings in language documentation \cite{thieberger2017ld}. Recent efforts have focused on leveraging foundation models like Whisper \cite{radford2023robust}, demonstrating that fine-tuned ASR models can be successfully integrated into language documentation workflows to generate full transcripts \cite{guillaume-etal-2022-fine, billings-mcdonnell-2025-connecting}.

Despite these advances, two critical barriers prevent the widespread adoption of ASR by field linguists. First, fine-tuning neural models requires technical expertise, specialised corpus formatting, and GPU management that fall outside the typical linguist's toolkit. Second, integrating ASR into a new documentation project introduces a ``cold start'' problem regarding data selection. Because manual annotation is highly resource-intensive, it is vital to train initial models as efficiently as possible. However, field linguists currently lack empirical guidelines on what type of recordings constitute the best seed data, whether to prioritise acoustic cleanliness (e.g., low background noise, no overlapping speakers) or linguistic richness (e.g., vocabulary diversity and repetition) when selecting initial data for fine-tuning.

In response to these challenges, we present a comprehensive methodology tailored to the realities of low-resource language documentation. To overcome the technical barrier, we introduce a practical, no-code ASR pipeline that handles the entire workflow: from annotations in ELAN (the standard tool for linguistic annotation \cite{wittenburg-etal-2006-elan}), through data preparation and model fine-tuning, to diarised transcription and integration back into ELAN for post-editing. To overcome the strategic barrier, we establish empirically tested data selection guidelines that optimise the human-in-the-loop transcription effort.

Our contributions are two-fold:
\begin{itemize}
 \item \textbf{Easper, A Portable Open-Source Workflow:} We release an open-source pipeline that integrates ELAN with ASR model fine-tuning, enabling linguists to train and deploy models without expert engineering support.\footnote{The code is available at \url{https://github.com/Aso-UniMelb/Easper}}
 \item \textbf{Transcription Prioritisation Strategies:} Using this pipeline, we simulate a transcription prioritisation scenario across three low-resource languages of Vanuatu. To ensure our evaluation spans the acoustic realities of field linguistics, our dataset contrasts recent, higher-quality recordings (Bislama) with archival fieldwork data (Nafsan and Nguna). We systematically compare five data selection strategies across these corpora, demonstrating that prioritising lexical richness and acoustic-phonetic repetition is more important for achieving higher accuracy than acoustic cleanliness in early-stage model adaptation.
\end{itemize}

\section{Related Work}
Large multilingual ASR models (e.g., Whisper \cite{radford2023robust}, XLS-R \cite{babu2021xls}, MMS \cite{pratap2024scaling}, Omnilingual \cite{omnilingualasrteam2025omnilingualasropensourcemultilingual}) show impressive zero-shot capabilities but require fine-tuning for under-resourced languages due to non-standardised orthographies and specific transcription preferences \cite{bird-2020-sparse}. While fine-tuning neural models now outperforms traditional, resource-intensive Kaldi-based systems \cite{povey2011kaldi} for language documentation \cite{jones-etal-2024-comparing, guillaume-etal-2022-fine}, the process remains technically challenging for field linguists without dedicated support for corpus formatting and GPU management.

Efforts to democratize ASR include Elpis \cite{foley2019elpis}, which provides a graphical interface for training Kaldi and neural models \cite{adams-etal-2021-user} directly from ELAN files. However, Elpis requires local GPU infrastructure and lacks support for fine-tuning modern language models. Furthermore, even as tools become more accessible, a strategic barrier remains: there is little empirical guidance on how linguists should select the initial data for fine-tuning. Our work addresses both gaps by providing a cloud-based, no-code fine-tuning pipeline paired with an empirical analysis of data selection strategies to guide active elicitation.

\section{Easper: The Portable ASR Workflow}
To enable iterative model improvement during fieldwork, we developed \textbf{Easper} (\textbf{E}LAN-integrated \textbf{A}utomatic \textbf{Spe}ech \textbf{R}ecogniser), a lightweight, open-source pipeline designed for linguists with limited technical resources. The workflow consists of three stages: Data Preparation, Fine-Tuning, and On-Device Inference.

\subsection{Data Preparation}

The first stage of the pipeline, implemented in the Easper Dataset Generator (Figure~\ref{fig:generator}), processes ELAN (.eaf) files and audio recordings to create training data for ASR fine-tuning. Using the Python library \texttt{pympi-ling} \cite{pympi-1.71}, the module extracts annotations and performs a series of validation checks designed to support systematic data cleaning.

Each annotation is evaluated against validation rules. Since ASR models work with short audio segments, annotations longer than 30 seconds are flagged for manual review. Such segments are truncated in models and can negatively affect performance. The module also detects overlapping annotations across tiers. All identified issues are summarised in a structured report for correction in ELAN.

In addition, to assess transcription quality, the module computes corpus-level statistics, including character and word frequency distributions. These help identify spelling inconsistencies, typographical errors, and unintended code-switching. Character bigram statistics are also generated to reveal structural patterns and guide future data collection.

After revision, the module exports the final dataset as 16 kHz mono WAV files together with a CSV file linking each audio segment to its transcription. The dataset is then ready for upload to Google Drive and ASR model fine-tuning in the next stage.

\subsection{ASR Model Fine-Tuning}

A major obstacle to the adoption of ASR in language documentation is the computational cost associated with model training and fine-tuning. To address this challenge, our workflow leverages cloud-based platforms such as Google Colab \cite{sukhdeve2023google}. This separates the heavy computational work of fine-tuning from the limitations of typical laptops.

Among the multilingual ASR models currently available, we recommend using OpenAI's \texttt{Whisper-small} (244M) and Meta's \texttt{XLS-R} (300M) in our pipeline. These models were chosen based on three criteria: (1) they can be fine-tuned using freely available cloud computing resources; (2) they have demonstrated strong performance in extremely low-resource language settings; and (3) they are compact enough to run efficiently offline on standard CPU-based computers.

This infrastructure is designed to support an iterative, human-in-the-loop workflow. By prioritising models with compact file sizes (less than 1 GB), we ensure that trained models remain highly portable. Linguists can download fine-tuned models from the cloud and deploy them directly within the offline Easper desktop application. This design enables the development and use of custom ASR systems in remote fieldwork settings, while minimising the need for advanced programming or machine learning expertise.

\subsection{On-Device Transcription}

\begin{figure}[t]
\centering
    \begin{subfigure}[b]{0.24\textwidth}
    \raggedleft
    \includegraphics[width=1\textwidth]{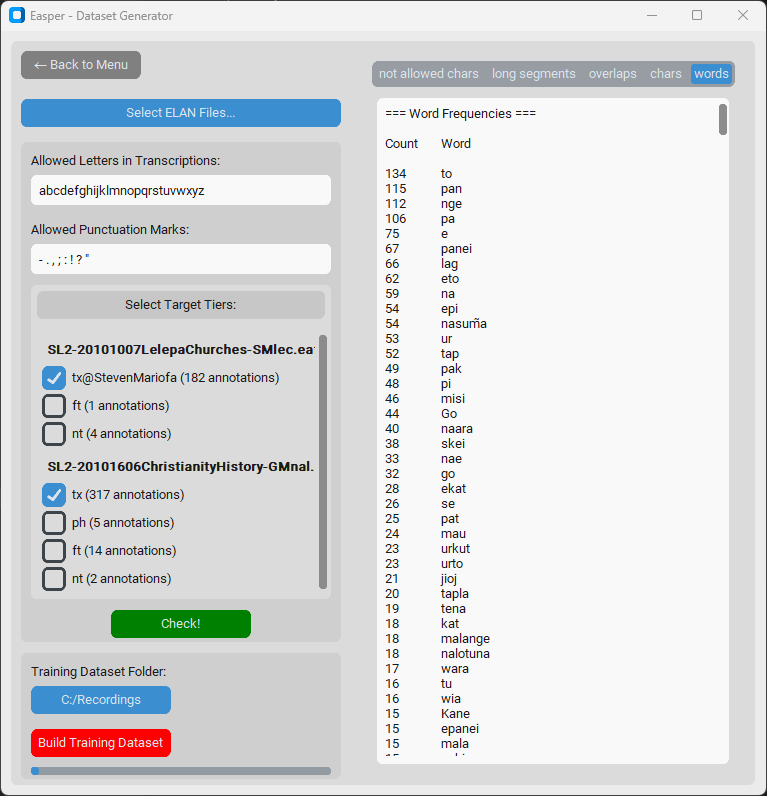}
    \caption{Dataset Generator}
    \label{fig:generator}
    \end{subfigure}
    \hspace{0.02\textwidth}
    \begin{subfigure}[b]{0.20\textwidth}
    \raggedright
    \includegraphics[width=0.9\textwidth]{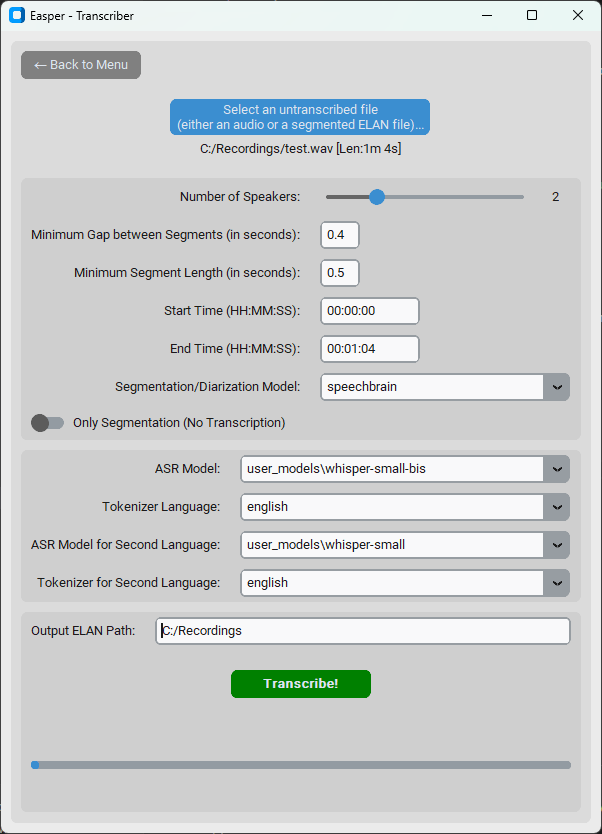}
    \caption{Transcriber}
    \label{fig:transcriber}
    \end{subfigure}
    \caption{Screenshot of Easper Desktop Application}
\end{figure}

The final stage applies the fine-tuned ASR model to new unannotated audio recordings through a four-step process: (1) speaker diarisation, (2) segmentation, (3) speech recognition, and (4) ELAN export.

For speaker diarisation and segmentation, simple silence detection and amplitude-based segmentation were tested, but these methods were unreliable in noisy recordings or with varied speaking styles. For robust segmentation, we use either the \texttt{SpeechBrain} toolkit \cite{ravanelli2021speechbrain} or \texttt{pyannote-audio} \cite{bredin2023pyannote}. These tools detect speaker boundaries and divide the audio into speaker-specific utterances. Since linguists know the number of speakers in advance, providing this information improves diarisation accuracy. The resulting segments are then transcribed using the fine-tuned Whisper model.

To ensure accessibility without technical overhead, the pipeline is packaged as a Python desktop application. This enables local, offline processing of sensitive field data. The graphical interface (Figure~\ref{fig:transcriber}) is straightforward: users select a trained model, input an audio file, and adjust a segmentation sensitivity slider to accommodate varying speaking paces and hesitations.

The transcription output is written back to a new ELAN file. Each speaker is assigned a separate tier, and each segment is aligned with its audio interval. This supports semi-automatic transcription, where the ASR provides a first draft, and the linguist performs revision. The modular design allows retraining and retranscription as more data becomes available.

\section{Methodology}
To guide transcription prioritisation during fieldwork, we propose a data selection methodology that evaluates the intrinsic acoustic and linguistic properties of unannotated audio. Unlike standard active learning paradigms that rely on model-uncertainty metrics over isolated speech segments, our approach is strictly tailored to the operational realities of language documentation. In fieldwork, manual transcription is conducted through continuous recording sessions, such as a specific interview, a folktale, or an elicitation task, rather than a randomised assortment of isolated utterances drawn from various days and speakers. To accurately mirror this workflow, we enforce a strict session-level constraint on our pipeline: we treat each recording session as an indivisible, discrete unit of data that must be evaluated and added to the training pool in its entirety.

\subsection{Session Feature Extraction}
For every session within a language's corpus, we extract a set of independent features representing its overall acoustic quality and linguistic richness:
\begin{itemize}
 \item \textbf{Acoustic Features}: Signal-to-Noise Ratio (SNR) and the proportion of overlapping speech (OVR).
 \item \textbf{Linguistic Features}:  For session $s$ we measure lexical diversity using the standard Type-Token Ratio,
\[
\text{TyTo}(s) = \frac{\text{\#Types}}{\text{\#Tokens}},
\]
 However, TyTo overly rewards words occurring only once, which are difficult for neural networks to learn without repetition, and it is insensitive to session duration. A length-sensitive alternative, the MATTR measure \cite{Covington01052010}, addresses the duration issue by averaging type-token ratio over a fixed-size sliding window of tokens. However, MATTR measures local lexical density within short windows, which is poorly suited to our highly variable and often short sessions; our objective is instead to capture session-level repetition. We therefore introduce the Normalised Token-to-Type Ratio,
\[
\text{ToTy}(s) = \frac{\text{\#Tokens}}{\text{\#Types} \times \text{Duration}},
\]
the average number of repetitions per unique word, normalised by session length so that longer sessions do not score higher.
\end{itemize}

\subsection{Data Selection Strategies}

To simulate a transcription prioritisation workflow, we incrementally add training data session-by-session. We define five distinct data selection strategies that rank the available unannotated sessions based on the extracted features:

\begin{itemize}
 \item \textbf{Baseline}: Sessions are added in a randomised order;
 \item \textbf{SNR Priority}: Sessions are added in descending order of Signal-to-Noise Ratio (cleanest background first);
 \item \textbf{Minimal Overlap Priority}: Sessions are added in ascending order of Speaker Overlap Rate (least overlapping speech first);
 \item \textbf{TyTo Priority}: Sessions are added in descending order of Type-Token Ratio (highest vocabulary diversity first);
 \item \textbf{ToTy Priority}: Sessions are added in descending order of Normalised Token-to-Type ratio (highest average repetition of words first).
\end{itemize}

\section{Experimental Setup}

To evaluate the effectiveness of our proposed methodology, we designed an experimental setup across three distinct fieldwork projects to assess how prioritising these different acoustic or linguistic characteristics impacts the data efficiency of the models.

\subsection{Dataset}
We utilise a corpus comprising ELAN-annotated field recordings from three languages spoken in Vanuatu (two Oceanic and one Creole), accessible with permission on the PARADISEC catalogue.\footnote{\url{https://paradisec.org.au/}}

\begin{itemize}
 \item \textbf{Bislama} (ISO 639-3: \texttt{bis}) is the national language of Vanuatu, spoken by approximately 300,000 people. It is a young, largely English-lexified, Creole that developed in the second half of the 19th century \cite{crowley2004bislama}.
 \item \textbf{Nafsan} (ISO 639-3: \texttt{erk}) is one of the 138 indigenous languages of Vanuatu and has around 5,000 speakers. The Nafsan writing system was originally created by missionaries in the 1860s and has been largely unchanged since then, except for the more recent addition of tildes to indicate co-articulated labio-velars, \~{p} and \~{m} \cite{thieberger2006grammar}.
 
 \item \textbf{Nguna} (ISO 639-3: \texttt{llp}), also known as North Efate, is spoken by approximately 9,500 people across the northern area of Efate and adjacent offshore islands. Its phonology and grammar were extensively documented in earlier descriptive works, making it a valuable comparative anchor in our dataset \cite{Facey1988Nguna}.
\end{itemize}

This selection offers diverse recording and annotation conditions. We contrast recent, high-quality audio (Bislama) with archival field recordings (Nafsan, Nguna), and highly accurate transcriptions (Bislama, Nguna) with noisier labels (Nafsan). Table~\ref{tab:dataset_stat} details the corpus statistics for these languages prior to fine-tuning, highlighting the varying scales of available data.

\begin{table}
\setlength{\tabcolsep}{3pt}
\centering
\footnotesize
\caption{Dataset statistics for the target languages.}
\label{tab:dataset_stat}
\begin{tabular}{lccrrr}
\toprule
Language & Total Audio & Sessions & Segments & Tokens & Types \\
\midrule
Bislama (bis) & 13h45m & 49 & 11,575 & 123,093 & 4,801 \\
Nafsan (erk)  & 14h50m & 32 & 9,245  & 107,775 & 8,526 \\

Nguna (llp)   & 01h01m & 7  & 1,098  & 7,976   & 852  \\
\bottomrule
\end{tabular}
\end{table}

\subsection{Implementation Details}

For each of the three languages, we initialise a separate instance of the pre-trained Whisper-Small foundation model (244M parameters). To rigorously evaluate the maximum potential of each data selection strategy, we perform full fine-tuning of the entire model architecture for 3 epochs per step, utilising a batch size of 8 and a learning rate of 1e-5. Prior to training, acoustic features are extracted by estimating the session SNR under the OM-LSA framework \cite{cohen2001speech} and detecting speaker overlaps via \texttt{pyannote} \cite{bredin2023pyannote}.

At each training step, we add the next highest-ranked session to the training pool, based on the chosen strategy. To account for large variance in recording lengths, we implement a minimum duration threshold; if a selected session is shorter than this threshold, we continue adding the subsequent highest-ranked sessions until the threshold is met for that training step. The model is then fine-tuned on the accumulated data.

\begin{figure*}[t]
  \centering
  \includegraphics[width=0.95\linewidth]{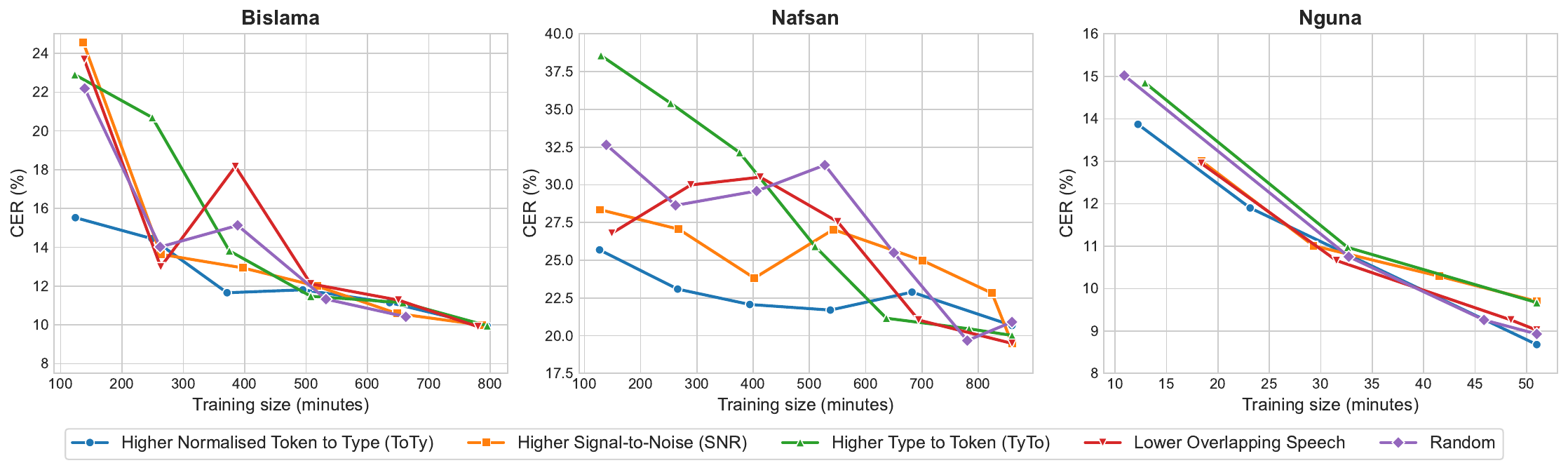}
  \caption{Learning curves demonstrating the impact of data selection strategies on Character Error Rate (CER).}
  \label{fig:learning_curves}
\end{figure*}

\subsection{Evaluation Metric}

We adopt Character Error Rate (CER) as our primary evaluation metric to better capture utility for post-editing by linguists. While Word Error Rate (WER) can be too harsh on non-standard spellings in low-resource languages, and Phoneme Error Rate (PER) complicates evaluation with language-specific symbol ambiguities \cite{adams-etal-2021-user}, CER directly approximates the keystrokes needed for post-editing. Defined as the minimum character edits required to convert the system output into the reference text, CER provides a fairer assessment of practical model performance.

\section{Results and Discussion}

We evaluated the five data selection strategies across our languages by incrementally fine-tuning the models session-by-session. To preserve sufficient training data for Nguna, given its extremely limited corpus (61 minutes total), its test set was scaled to 10 minutes, while a fixed 30-minute test set was maintained for Bislama and Nafsan.

\subsection{Impact of Selection Strategies on Learning Curves}
As illustrated in Figure \ref{fig:learning_curves}, while all strategies generally decrease the CER as more transcribed data is added, their early-stage learning trajectories diverge significantly. The spikes observed in some trajectories likely reflect catastrophic forgetting induced by small, heterogeneous training increments. The Normalised Token-to-Type Ratio (ToTy) priority tends to outperform the other strategies during these early stages, achieving the lowest CER across the target languages.

\subsection{Robustness of Foundation Models vs. Lexical Sparsity}

The results of these simulated transcription prioritisation experiments challenge conventional assumptions about fieldwork data collection. A primary finding of this study is that linguistic distribution outranks acoustic cleanliness when bootstrapping an ASR model for a new language.

The mediocre performance of the SNR and Minimal Overlap strategies suggests that massive pre-trained foundation models like Whisper are already highly robust to the non-stationary background noises and overlaps typical of field recordings. What the foundation model lacks is not acoustic robustness, but rather the specific vocabulary and orthographic rules of the target language. By explicitly feeding the model sessions with high lexical density early in the training process, it rapidly aligns its existing acoustic representations to the novel linguistic space.

\subsection{Lexical Breadth and Depth}
Furthermore, by comparing the Type-Token Ratio (TyTo) with our Normalised Token-to-Type Ratio (ToTy), we observed a critical tension between vocabulary breadth and depth.

While TyTo exposes the model to a rich vocabulary, prioritising ToTy tends to perform better. This confirms that neural networks require sufficient acoustic-phonetic repetitions to reliably learn new grapheme-to-phoneme mappings. A session with a massive but sparse vocabulary (high TyTo) is practically less useful for early fine-tuning than a session that deeply reinforces a core set of vocabulary through repetition (high ToTy).

\subsection{Practical Recommendations for Linguists}

Based on our pipeline development and empirical evaluations, we offer the following guidelines to optimise fieldwork for ASR integration. While capturing clean audio using directional microphones in quiet settings remains vital for archival integrity, ASR fine-tuning benefits most from linguistic richness. Linguists should actively elicit lexically dense, repetitive narratives and prioritise their transcription, even if the recording contains sub-optimal background noise.\footnote{For practical guidelines on ELAN annotation formatting compatible with the Easper pipeline, see \url{https://github.com/Aso-UniMelb/Easper}}

\section{Conclusion}
 
We have presented Easper, a practical, open-source ASR pipeline tailored for language documentation. By integrating ELAN, Whisper, and cloud computing, the workflow enables field linguists to independently handle data preparation, model fine-tuning, and diarised transcription without requiring programming expertise or specialised hardware.

Beyond providing this infrastructure, we addressed the strategic cold start problem of ASR bootstrapping through simulated transcription prioritisation across three Vanuatu languages. Crucially, our findings challenge the assumption that clean audio is paramount for early-stage training. Instead, we demonstrate that prioritising linguistic richness, specifically lexical breadth and acoustic-phonetic repetition, is significantly more effective than acoustic cleanliness for accelerating model adaptation. These results offer field linguists empirical guidelines for optimising their transcription efforts.

Future work will adapt the Easper pipeline for massive legacy archives, such as PARADISEC, which contains over 21,000 hours of audio spanning 1,400 languages, with many lacking transcripts or metadata. Implementing Easper's iterative workflow is expected to offer a crucial means of accessing and utilising this extensive and previously inaccessible material.

\section{Acknowledgments}
This research was supported by The University of Melbourne's Research Computing Services and the Petascale Campus Initiative. Funding for Thieberger's work and for PARADISEC is from ARC DP250102214 and the Language Data Commons of Australia.  

\section{Generative AI Use Disclosure}
In accordance with ISCA policy, the authors disclose the use of Generative AI tools to assist in editing and polishing the language of this manuscript. These tools were not used to produce any significant portion of the scientific content, experimental design, or core ideas. The authors take full responsibility and accountability for the final content of this paper.

\bibliographystyle{IEEEtran}
\bibliography{mybib}

\end{document}